\documentclass{article}
\usepackage{amsmath}
\usepackage{subcaption}
\PassOptionsToPackage{numbers, compress}{natbib}

\usepackage[nonatbib,preprint]{neurips_2024}
\usepackage{amsmath}

\usepackage[utf8]{inputenc} 
\usepackage[T1]{fontenc}    
\usepackage{url}            
\usepackage{booktabs}       
\usepackage{amsfonts}       
\usepackage{nicefrac}       
\usepackage{microtype}      
\usepackage{algorithm}
\usepackage{algorithmicx, algpseudocode}
\usepackage[utf8]{inputenc} 
\usepackage[T1]{fontenc}    
\usepackage[bookmarks=false]{hyperref}       
\usepackage{url}            
\usepackage{booktabs}       
\usepackage{amsfonts}       
\usepackage{nicefrac}       
\usepackage{microtype}      
\usepackage{xcolor}         
\usepackage{caption}
\usepackage{graphicx}
\usepackage{bm,multirow}
\usepackage{multicol}
\usepackage{adjustbox}
\usepackage{bm}
\usepackage{wrapfig}

\def\eqref#1{(\ref{#1})}
\def\mQ{{{Q}}}
\def\mK{{{K}}}
\def\mI{{{I}}}
\def\mV{{{V}}}
\def\vf{{{f}}}
\def\vz{{{z}}}

\usepackage{amsmath,amsfonts,bm}

\def\eqref#1{equation~\ref{#1}}

\def\1{\bm{1}}

\def\vf{{\bm{f}}}

\def\vz{{\bm{z}}}

\def\mI{{\bm{I}}}

\def\mK{{\bm{K}}}

\def\mQ{{\bm{Q}}}

\def\mV{{\bm{V}}}

\DeclareMathAlphabet{\mathsfit}{\encodingdefault}{\sfdefault}{m}{sl}
\SetMathAlphabet{\mathsfit}{bold}{\encodingdefault}{\sfdefault}{bx}{n}

\title{Unified Editing of Panorama, 3D Scenes, and Videos Through Disentangled Self-Attention Injection }

\author{ Gihyun Kwon$^{2}$,\quad Jangho Park$^3$, \quad Jong Chul Ye$^{1,2,3}$\\
Kim Jaechul Graduate School of AI$^1$, Department of Bio and Brain Engineering$^2$,\\
 Robotics Program$^3$, KAIST \\
\texttt{\{cyclomon,jhq1234,jong.ye\}@kaist.ac.kr} \\
}

\begin{document}

\maketitle

\begin{abstract}
 While text-to-image models have achieved impressive capabilities in image generation and editing, their application across various modalities often necessitates training separate models. 
 Inspired by existing method of single image editing with self attention injection and video editing with shared attention, we propose a novel unified editing framework that combines the strengths of both approaches 
by utilizing only a basic 2D image text-to-image (T2I) diffusion model. 
  Specifically, we design a sampling method that facilitates editing consecutive images while maintaining semantic consistency utilizing shared self-attention features during both reference and consecutive image sampling processes. Experimental
  results confirm that our method enables editing across diverse modalities including 3D scenes, videos, and panorama images. 
\end{abstract}

\section{Introduction}
Recent text-to-image (T2I) models have demonstrated impressive image generation capabilities, and their applications have been extended to editing attributes or textures of real images~\cite{pnp,p2p,mokady2022nulltext}. Additionally, there have been efforts to extend diffusion models beyond the 2D image domain, including 3D scene or object generation~\cite{dreamfusion,prolificdreamer} and text-to-video generation~\cite{show1,lumiere,magicvideo}. Despite these advancements, applying diffusion models to different modalities requires training separate models for each modality. These fragmented models result in increased difficulty in attribute editing and higher resource consumption. To address these challenges, we propose a novel editing method that enables unified editing across all modalities using only a basic 2D image text-to-image diffusion model. 

Our approach leverages the inherent sequential nature of images in different modalities: 3D scenes reconstructed from multi-view images, videos comprised of frames, and high-resolution panoramas stitched from sub-images.
Drawing inspiration from existing methods such as self-attention injection (PNP)\cite{pnp} for single image editing and shared attention for video editing (Pix2Vid)\cite{pix2vid}, we introduce a unified editing framework. This framework incorporates the strengths of both approaches, facilitating editing of consecutive images while ensuring semantic consistency across the images.

\begin{figure}[t!]
    \centering
   \includegraphics[width=1.0\linewidth]{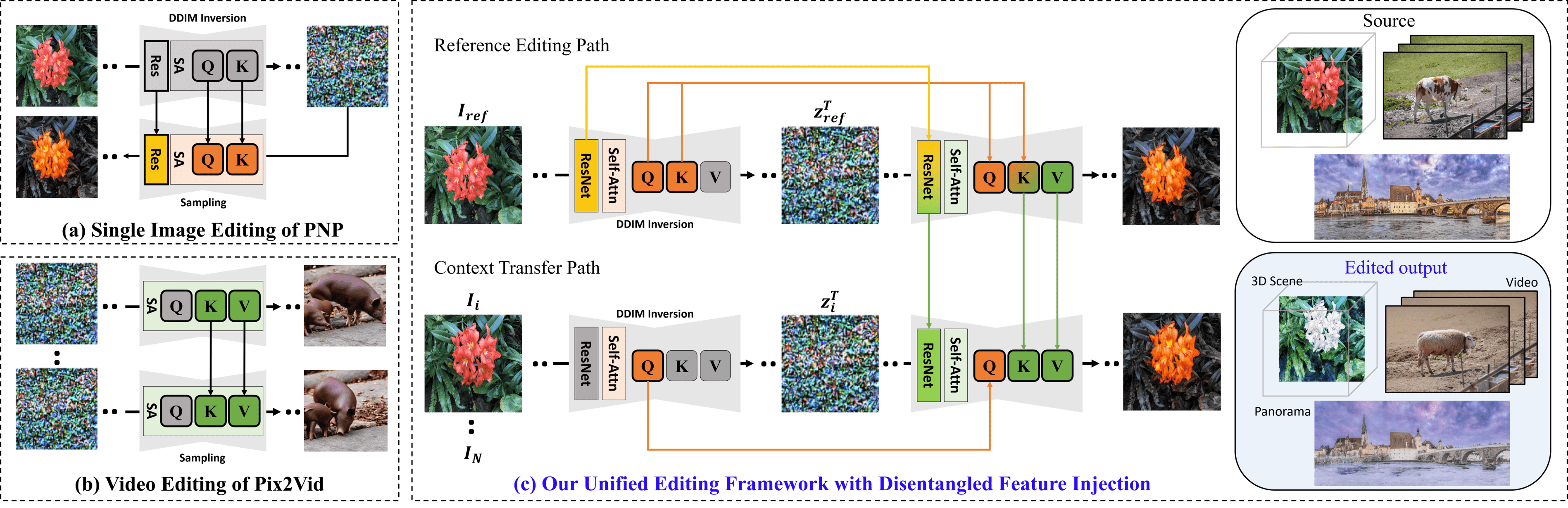}
    \caption{\textbf{Method Overview.} (a) Plug-and-play diffusion \cite{pnp} single image editing. The method inject resnet, query, key features during sampling process for disentangled image editing. (b) Pix2Vid \cite{pix2vid} proposes to propagate context from one sampling path to consequent sampling with injecting key and value features. (c) Our proposed method apply DDIM inversion to series of images to obtain the initial noise. During inversion, we extracted the resnet and self-attention features from Diffusion U-Net. Starting from initial inverted noise, we sample the outputs with feature injection. We inject inverted resnet and self-attention features to image editing path. For consistent editing, we propagate key and value features of edited sample to consequent sampling path. Our method enable editing on various modalities including 3D scene, panorama, and video. 
}
    \label{fig:fig_main}
    \end{figure}

More specifically, as illustrated in Figure \ref{fig:fig_main}(a), PNP diffusion method leverages self-attention features extracted during the inversion step to maintain the structural information of the source image while editing targeted attributes. The resnet feature, query feature, and key feature of self-attention layer are injected during the sampling stage, enabling a disentangled editing approach. In Pix2Vid of Figure \ref{fig:fig_main}(b), the method try to edit videos using only 2D image diffusion. During the sampling stage, self-attention's key and value features are shared between two sampling paths. This ensures that both outputs share the same context, resulting in consistent edits across the video frames. 

Our framework combines the strengths of both existing methods to achieve unified editing across modalities. As depicted in Figure \ref{fig:fig_main}(c), we employ two parallel paths: disentangled editing on reference image using inverted self-attention features and context transfer using shared self-attention features. The unified framework of editing and context transfer simultaneously ensures superior editing performance and cross-image consistency. To further enhance performance, we introduce a scheduling mechanism that dynamically adjusts the injection strength based on the timestep.


Our main contributions are as follows:

\begin{itemize}
    \item We propose a novel unified editing method that enables seamless editing across panorama images, videos, and 3D scenes only using a single 2D image text-to-image diffusion model.
    \item We design a simple and powerful training-free sampling method for achieving both of disentangled editing and cross-image context consistency with unified framework of single-image editing and sequential image editing methods.
    \item Our method can be easily extended to various applications such as editing with custom concept and localized editing.

\end{itemize}

\section{Related Works}
\noindent\textbf{2D Image Editing.} The development of diffusion models has revolutionized real image editing techniques. Early approaches focused on text-guided image style transfer and attribute editing using CLIP models and unconditional diffusion models~\cite{diffusionclip,diffuseit,asyrp}. As text-to-image diffusion models advance, various editing techniques have been developed, particularly based on the open-source Stable Diffusion model~\cite{ldm}. These include image translation~\cite{pnp}, attention-based editing~\cite{pnp,mokady2022nulltext}, and guided sampling-based editing~\cite{pix2pixzero}. Additionally, several researches explored methods~\cite{csd,syncdiffusion,multidiffusion} for editing and generating large images beyond fixed resolutions.

\noindent\textbf{3D Editing.} Pre-trained NeRF~\cite{nerf} models are often used for editing 3D scenes and objects. Early methods ~\cite{clipnerf,nerfart,blending} applied CLIP~\cite{clip} models for text-guided editing of NeRF attributes or styles, but these were limited by CLIP's performance. Recently, several approaches~\cite{vox-e,ednerf,cds,csd} tried to transfer the generative prior of diffusion models to NeRF using Score Distillation Sampling (SDS)~\cite{dreamfusion} and its variations. However, SDS sampling methods face limitations in stability and overall performance compared to diffusion model reverse sampling. Recently, more advanced 3D reconstruction method of Gaussian splatting~\cite{gaussian} enabled faster and stable reconstruction, also contributed to the improvement of editing performance~\cite{gaussianeditor}. 

\noindent\textbf{Video Editing.} Early video editing approaches adapted 2D text-to-image diffusion models by adding temporal attention layers~\cite{tuneavideo}, enabling continuous editing without requiring large video data. However, these methods were limited by the lack of video generative priors. To address this, researchers have trained text-to-video models on large video datasets, achieving remarkable performance~\cite{magicvideo,show1,lumiere}. Various video editing methods~\cite{gen1,vmc} leveraged these text-to-video models with modifying the attention architecture to transfer the motion from source to output videos. However, video diffusion models also have limitation as it require significant computational resources.
To overcome this limitation, several methods ~\cite{fatezero,pix2vid} proposed to use attention feature modifications in 2D diffusion model sampling for editing sequential video frames. These methods achieve comparable performance to video diffusion models but some methods still require substantial resources for gradient computation or suffer from limited editing performance.

\section{Method}

In our approach, we propose an improved method by amalgamating single image editing method using self-attention injection (PNP) and sequential video editing method using shared attention framework (Pix2Vid) in order to leverage two advantages of superior single-image editing performance with consistent generation performance of video editing method. This combined approach enhances overall editing performance and maintains consistency across sequential images, making it applicable to editing 3D, video, and panorama images.

\subsection{Key Idea}
We first assume that there already exists a series of images ${I_1,I_2,...I_N}$ to be edited. These images can be frame images extracted from a video or multi-view images with known view direction and camera parameters for 3D scene data. For panorama image editing, a high-resolution real image is cropped into multiple images with a lower resolution (e.g., 512x512). Our goal is to semantically edit all these images so that they align with a given target text $t_{trg}$. At the same time, we aim to sample the images to ensure that the context, such as color and texture, is consistent between the output images. To achieve this, we will set one image from the series of images as a reference $I_{ref}$ and propagate the editing process of this image to the other images.

The U-Net architecture of text-to-image diffusion models consists of residual networks, self-attention, and cross-attention. 
As revealed in the previous studies~\cite{pnp}, the self-attention layer's query and key features are known to determine the overall structure and form of the image. Additionally, residual layer features, especially features in low-resolution layers, have been shown in previous studies to contain implicit information that determines the overall appearance of the image. Therefore, as shown in Figure~\ref{fig:fig_main}(a), the method proposed to extract self-attention and resnet features from DDIM inversion path then injected the corresponding features during sampling in order to edit the disentangled attribute without harming the structure of source images.

Based on this information, the previous study of Pix2Vid~\cite{pix2vid} (Figure \ref{fig:fig_main}(b)) has proposed a method to maintain context by using shared key and value feature from the self-attention layer in sampling paths. This method gives inspiration that key and value feature contains overall context of images. 
To address this, we devised a sampling method that performs editing and context preservation simultaneously by unifying the characteristics of the self-attention layer that have already been revealed. 

Specifically, for reference image editing, we employ the PNP diffusion technique to inject only resnet, query, and key features. This approach preserves the structural information of original image while achieving text-guided image editing performance. Simultaneously, during sequential image sampling, we utilize key and value features shared with the reference editing process to maintain context consistency. Additionally, we share the resnet feature to further ensure context preservation across images. More technical details are as follows.

\subsection{Algorithmic Details}


\noindent\textbf{Inversion and Feature Extraction}.
First, we send the reference image $I_{ref}$ and the $i$th image in the series $I_{i}$ to the noisy latent space through DDIM Inversion~\cite{ddim}. In this process, we extract the intermediate layer features of the diffusion U-net $\epsilon_\theta$. 

Specifically,   as shown in Figure \ref{fig:fig_main}(c),  during the sampling step $t$, for the reference image we extract the $l$-th layer resnet feature $\vf^{l,t}_{ref}$, and self-attention layer query and key features $\mQ^{l,t}_{ref},\mK^{l,t}_{ref}$ of the reference image $I_{ref}$, 
 which is calculated as:
\begin{equation*}
\mQ^{l,t}_{ref}=W^q\vf^{l,t}_{ref},\quad \mK^{l,t}_{ref}=W^k\vf^{l,t}_{ref} 
\end{equation*}
where  $W^q,W^k$ are weight parameters for projecting features to self attention. 
Similarly, for the $i$th image in the series $I_{i}$,  the intermediate feature $\vf^{l,t}_{i}$ are similarly extracted, from which we can obtain corresponding features $\mQ^{l,t}_{i}$ with replacing $\vf^{l,t}_{ref}$ into  $\vf^{l,t}_{i}$.
Following the settings from previous work\cite{pnp}, in $\mQ^{l,t}_{ref},\mK^{l,t}_{ref}$, and $\mQ^{l,t}_{i}$, $l$ is set as U-Net upsample layers $l=[4,7,9]$, and for $\vf^{l,t}_{ref}$, we set $l=4$ which is the lowest resolution of upsample layers.  
In order to prevent confusion, we notate the `inverted' features in this step as $\hat{\vf}_{ref}$,$\hat{\mQ}_{ref},\hat{\mK}_{ref}$,$\hat{\mQ}_{i}$

\noindent\textbf{Disentangled self-attention Injection}.
Starting from the noisy latent $z^T_{ref},z^T_{i}$ obtained from the previous inversion processes of $I_{ref}$ and $I_i$, respectively, we begin reverse diffusion process. During the sampling process, we aim to inject the features obtained in the previous steps appropriately to achieve image editing while maintaining the structure of the reference image,  and to propagate this editing path to other frames. The editing process of the reference image is inspired by plug-and-play diffusion features~\cite{pnp}, which is to inject the residual layer features and self-attention query, key features during reverse sampling. In this way, we can perform disentangled sampling for the target text while maintaining the core structure information of the reference image.

In the reverse sampling process using diffusion U-net $\epsilon_\theta$, we inject the obtained self-attention and resnet features as shown in Figure \ref{fig:fig_main}(c). In the editing path for the reference image $I_{ref}$, the noise prediction output of the U-Net is defined as follows:
\begin{eqnarray}\label{eq:edit}
    \epsilon^t_{ref} &=& \epsilon_{\theta}(\vz^t_{ref},t,c;[\hat{\mQ}_{ref},\hat{\mK}_{ref},\mV_{ref},\hat{\vf}_{ref}]),
    \end{eqnarray}
where $c$ is the text condition, $\mV_{ref}$ is value feature calculated from current resnet feature. In this process, query and key features from DDIM inversion $\hat{\mQ}_{ref},\hat{\mK}_{ref}$ maintain the structure information of the reference image $I_{ref}$ in the editing process starting from $z_{ref}$, and also injected residual layer feature $\hat{\vf}_{ref}$ prevents the semantic from changing excessively in the editing process.

To transfer the context information of the reference image to the sequential images $I_{i}$, we passed the key and value features $\hat{\mK}_{ref},\hat{\mV}_{ref}$ from the reference image sampling process to the generation path of $I_{i}$, and also passed the resnet feature $\hat{\vf}_{ref}$. However, injecting the above features may transfer excessive information about the reference $\mI_{ref}$, and the structure information of the original image $I_{i}$ is lost. To prevent this, we injected the query feature $\hat{\mQ}_{i}$ obtained from the inversion process of the image so that it can harness the structural information of $I_{i}$ (see Figure \ref{fig:fig_main}(c) bottom). This leads to the following:
\begin{eqnarray}\label{eq:editI}
    \epsilon^t_{i} &=& \epsilon_{\theta}(\vz^t_{i},t,c;[\hat{\mQ}_{i},\hat{\mK}_{ref},\mV_{ref},\hat{\vf}_{ref}]). 
\end{eqnarray}

\noindent\textbf{Injection Scheduling}. The use of feature injection as described above enables consistent image editing, but we found that the fidelity of the generated images is significantly reduced when it is applied to the entire timesteps. Since our framework separates the editing strength and context transfer components, we were able to generate high-quality images by varying the amount of self-attention injection according to the timesteps, thereby controlling the editing strength. Additionally, this timestep-variant injection strategy is essential because the degree of consistency required varies depending on the modality.

Thus, we define $t_{edit}$ as the step that is involved in editing, and $t_{context}$ as the step that is involved in context consistency. During the timestep $t_{edit}<t<T$, the sampling step should consider both of structure-maintained editing and context consistency. Therefore, we use noise estimation with above attention injected network of Eq. \eqref{eq:edit},\eqref{eq:editI}.
After $t_{edit}$, the editing-related branch does not need to be used during the sampling process. Therefore, instead of using the features,  $\hat{\mQ}_{ref},\hat{\mK}_{ref}$, $\hat{\vf}_{ref}$  and $\hat{\mQ}_{i}$,  extracted from the inversion step that were used to maintain the structure, we pass $\mK$, $\mV$, and $\vf$ calculated from the reference path itself to maintain only the context information between the images. Therefore, when $t_{context}<t<t_{edit}$, we use the following noise estimation output:
\begin{eqnarray}\label{eq:edit2}
    \epsilon^t_{ref} &=& \epsilon_{\theta}(\vz^t_{ref},t,c;[{\mQ}_{ref},{\mK}_{ref},\mV_{ref},{\vf}_{ref}]), \notag \\
    \epsilon^t_{i} &=& \epsilon_{\theta}(\vz^t_{i},t,c;[{\mQ}_{i},{\mK}_{ref},\mV_{ref},{\vf}_{ref}]). 
\end{eqnarray}
After $t_{context}$, we just use estimated output without using any self-attention injection strategy.

\begin{figure}[t!]
    \centering
    \includegraphics[width=0.9\linewidth]{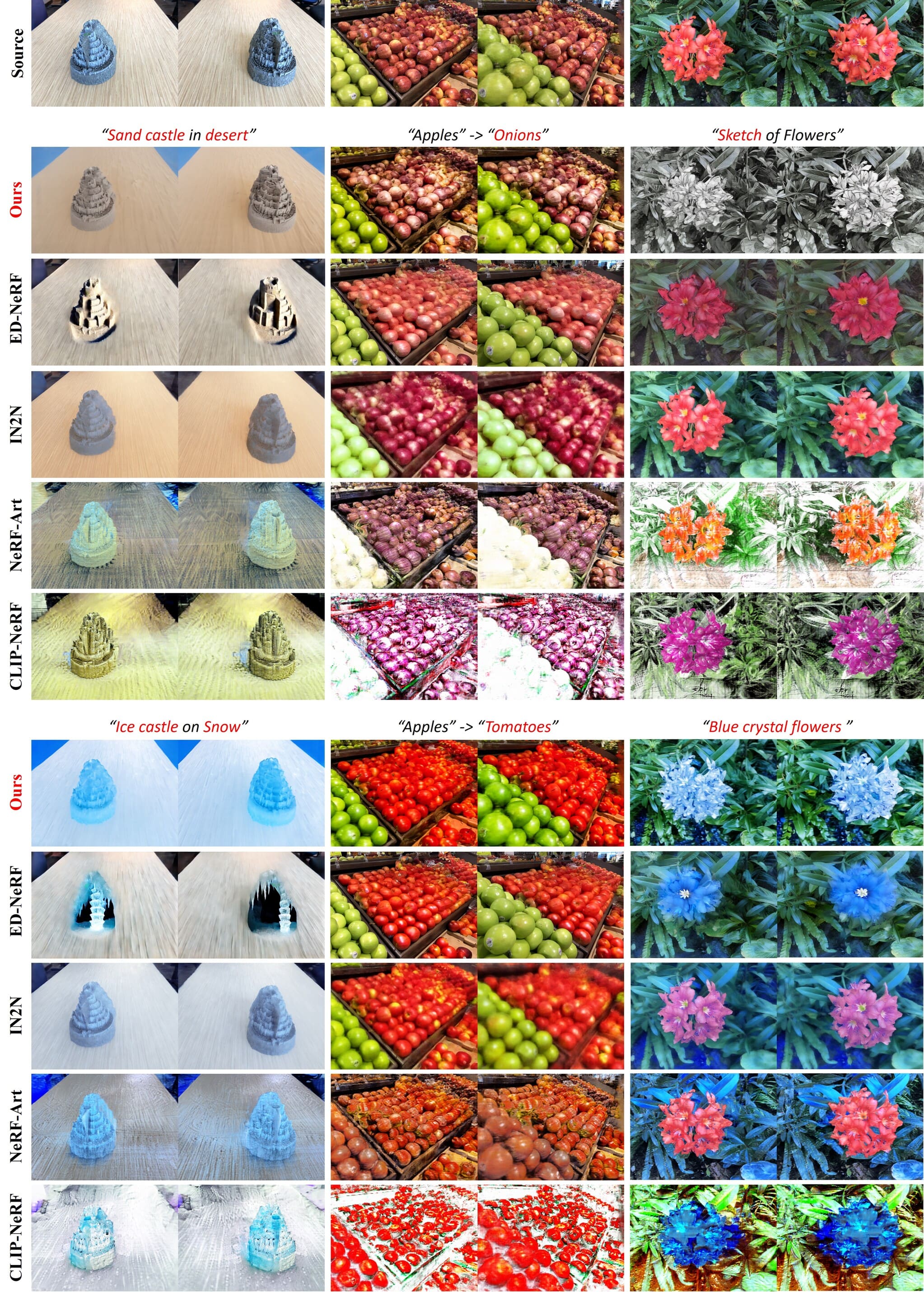}
    \caption{\textbf{Qualitative Evaluation of 3D scene editing}. 
    Our method outperforms baselines in both of semantic editing and overall style transfer.
}
    \label{fig:fig_nerf}
     \vspace{-0.4cm}
    \end{figure}

\begin{figure}[!t]
    \centering
    \includegraphics[width=\linewidth]{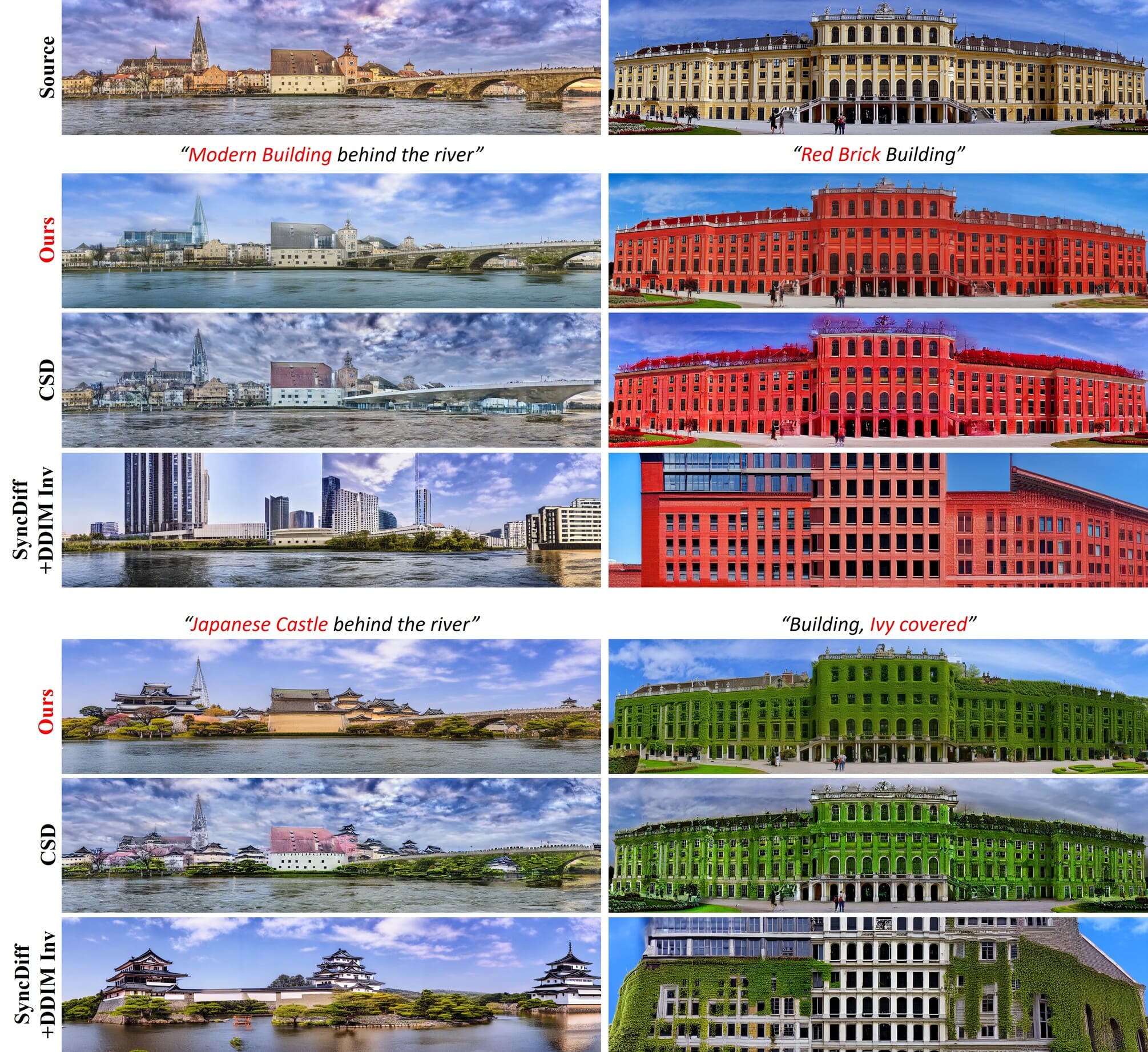}
    \caption{\textbf{Qualitative Evaluation of Panorama editing}. We compare the sampled panorama outputs. Our method outperforms baselines showing realistic output with high structural consistency.
}
    \label{fig:fig_panorama}
     \vspace{-0.4cm}
    \end{figure}

\section{Experiments}
\vspace{-0.2cm}
\subsection{3D Scene Editing}
\vspace{-0.2cm}
\noindent\textbf{Baselines. }For evaluating 3D scene editing, we compared our proposed method with state-of-the-art 3D scene editing methods. For early method which use pre-trained CLIP, we used CLIP-NeRF~\cite{clipnerf} and Nerf-Art~\cite{nerfart} as our baseline models. They fine-tuned the pre-trained model with guiding rendered images with text conditions using CLIP. For fair comparison, both models were implemented with same TensoRF\cite{tensorf} backbone. For improved baselines which utilize text-to-image diffusion model, we set Instruct-NeRF2NeRF(IN2N)~\cite{in2n} and ED-NeRF~\cite{ednerf} as our baselines. IN2N is an image editing method that resembles our approach in its utilization of a T2I  model to edit rendered images. It sequentially replaces training images with edited images during the NeRF training process to achieve the desired editing. ED-NeRF is a more recent image editing technique that employs a score distillation method to incorporate the generative prior of a diffusion model into NeRF. It first trains NeRF in latent space and then fine-tunes the parameters using score distillation. 

\noindent\textbf{Qualitative Results. } To compare the qualitative editing results between our method and existing approaches, we present the sampling results in Figure \ref{fig:fig_nerf}. We evaluated text-based 3D scene editing on three different scenes. Early models of CLIP-NeRF can modify the semantics of the scene based on the text description but suffer from severe artifacts. Advanced model of NeRF-Art exhibit reduced artifacts compared to early models but still the outputs deviate from the target text description and introduce unwanted artifacts. This is attributed to the direct transfer of CLIP information to NeRF, due to the low performance of CLIP as it is not designed for generation task. More recent method of IN2N produces fewer artifacts but fails to fully adhere to the text description, often remaining at the original scene stage. This is due to the performance limitations of InstructPix2Pix, which is the main editing model used in IN2N. Additionally, IN2N's inconsistent image editing approach, where each training image is edited independently, prevents the rapid propagation of individual image editing outputs to NeRF. In case of ED-NeRF, it successfully modifies the scene semantics according to the text description but significantly damages the structure of the source scene. This artifact is likely caused by the stochastic nature of score distillation, leading to unstable training. Our proposed method effectively edit the scene while preserving the structure of the source scene and following the text description. It outperforms the baseline models not only in semantic object editing but also in overall style transfer. 

\noindent\textbf{Quantitative Results. } 
To quantitatively evaluate the editing performance, we measured CLIP directional score~\cite{stylegan-nada} and CLIP image context consistency score. To evaluate the text-output semantic alignment, we calculated CLIP scores. However, for detailed measurement of editing performance, we calculated improved directional CLIP score which considers both of source and output scene and text descriptions. Also, we calculated CLIP image consistency score which measures CLIP similarity score between the individual edited outputs within the series of outputs. With this method, we can measure whether the output is semantically consistent. For validation, we calculated scores using 6 different scenes edited with 2 text condition for each scene. For each scene, we rendered 120 images from novel camera view directions and calculated averaged scores for those images. For CLIP model, we used CLIP-L/14 which is large model trained with 14x14 patch.

        
			
    
				
				
				
				
				
				

In Table \ref{table:nerf}, early model of CLIPNerf shows high text score with decent view consistency, but we already observed severe artifact in our qualitative results. For other methods of Nerf-Art, IN2N, and ED-NeRF, the methods failed to follow the text condition showing relative low score in directional CLIP score. Also, for ED-NeRF, the model suffer from low cross-view consistency score. We conjecture that it comes from unstable training nature of score distillation sampling. In our proposed method, we obtained the best text-image alignment score and second best in cross-view context consistency scores. These results indicate that our method can consistently edit the source scenes with target conditioning text. Rendered samples are in our anonymous project page.\footnote{\url{https://unifyediting.github.io/}}

\begin{figure}[t!]
    \centering
    \includegraphics[width=\linewidth]{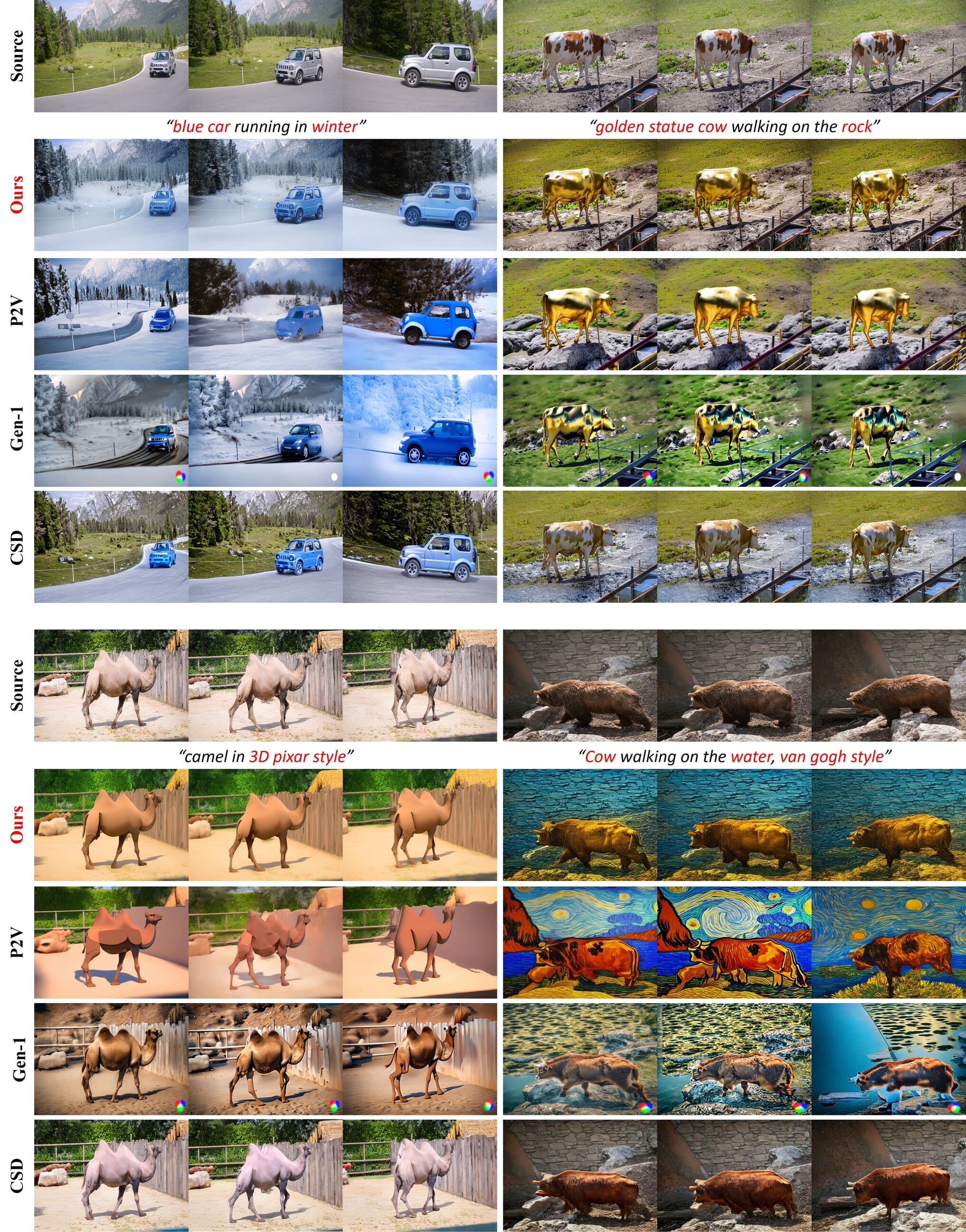}
    \caption{\textbf{Qualitative evaluation of video editing}. Our method shows better cross-frame consistency with text-output semantic alignment compared to baseline methods.
}
    \label{fig:fig_video}
    \vspace{-0.4cm}
    \end{figure}

\begin{figure}[t!]
    \centering
    \includegraphics[width=1.0\linewidth]{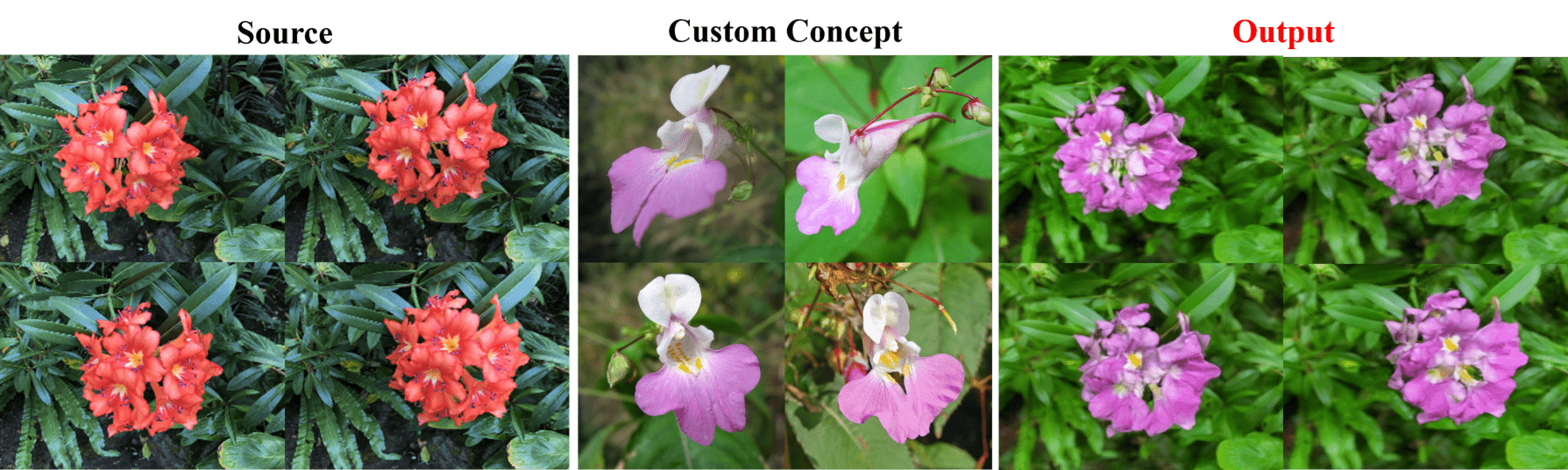}
    \caption{\textbf{Results of custom concept editing.} We can successfully transfer the semantic of custom concepts to various modalities with our proposed method. 
}
    \label{fig:fig_custom}
     \vspace{-0.4cm}
    \end{figure}

\begin{figure}[t!]
    \centering
    \includegraphics[width=1.0\linewidth]{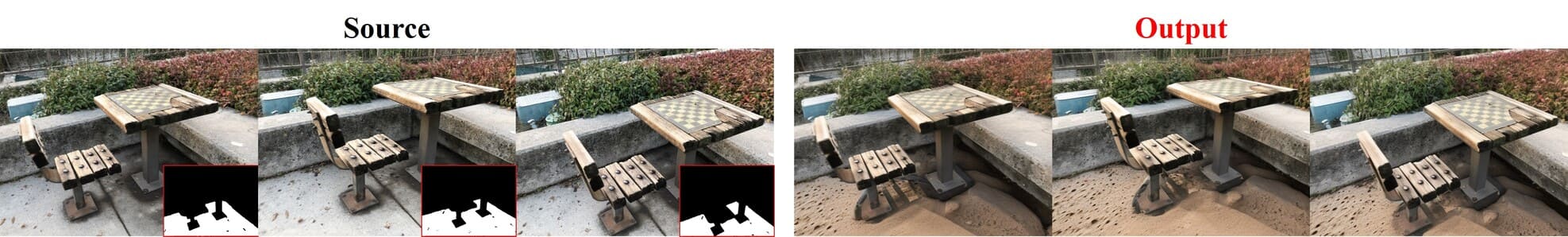}
    \caption{\textbf{Results of local editing.} We show the edited output using text condition \textit{"Table and chair on the desert"}. Our method can independently edit the designated region with additional masks. 
}
    \label{fig:fig_mask}
    \vspace{-0.4cm}
    \end{figure}

\vspace{-0.1cm}    
\subsection{Panorama Image Editing}
\vspace{-0.2cm}
\noindent\textbf{Baselines. } For panorama image editing, we leveraged two different baseline methods, SyncDiffusion~\cite{syncdiffusion} and Collaborative Score Distillation (CSD)~\cite{csd}. SyncDiffusion uses posterior sampling based method for maintaining the context of cross-patch sampling path. Also, CSD use score distillation sampling to edit the large resolution and video images with cross-image consistent optimization. As our baseline of SyncDiffusion is originally designed for novel image generation, we combined the method with DDIM inversion to apply the method in image editing.

\noindent\textbf{Qualitative Results. } In Figure \ref{fig:fig_panorama}, we show the qualitative comparison results. In SyncDiffusion output, we can observe that the output image follows text description with realistic quality, but the image failed to maintain the original source image structure. In CSD output, the output can edit the source images with target text with decent consistency. However, the method still could not fully edit the image with target text, and often the method suffer from unwanted artifact such as jittering in background regions. In our method, we can successfully change the semantic of source image to target text, while maintaining the original structure of source images.

\begin{table}[!t]
\begin{minipage}{0.31\textwidth}
    \centering
    \resizebox{1.0\textwidth}{!}{
    			\begin{tabular}{@{\extracolsep{5pt}}ccc@{}}
				\hline
				\multirow{2}{*}{\textbf{Method}}  & \multicolumn{2}{c}{\textbf{CLIP score}} \\
    
				
				\cline{2-3} 
				
				
				& Text sim$\uparrow$& Consistency$\uparrow$ \\
				
				\hline
				CLIP-NeRF &0.2170&0.9433\\
				NeRF-Art  &0.1472&0.9512\\
				IN2N  &0.1713&0.9394\\
				ED-NeRF  &0.1835&0.9091\\
                \hline
				
				\textbf{Ours}  & {0.2351}&{0.9480}\\
				\hline
				
			\end{tabular}
   
   }\subcaption{3D scene editing}\label{table:nerf}
\end{minipage}
\hfill
\begin{minipage}{0.34\textwidth}
    \centering
    \resizebox{1.0\textwidth}{!}{
    \begin{tabular}{@{\extracolsep{5pt}}ccc@{}}
				\hline
				\multirow{2}{*}{\textbf{Method}}  & \multicolumn{2}{c}{\textbf{Score}} \\
    
				
				\cline{2-3} 
				
				
				& Text sim$\uparrow$& Structure$\uparrow$ \\
				
				\hline
				SyncDiff  &0.0942&0.3133\\
				CSD  &0.1812&0.6232\\
                \hline
				
				\textbf{Ours}  & {0.2053}&{0.5725}\\
				\hline
				
			\end{tabular}}
   \subcaption{Panorama editing}\label{table:panorama}
\end{minipage}
\hfill
\begin{minipage}{0.32\textwidth}
    \centering
    \resizebox{1.0\textwidth}{!}{
    \begin{tabular}{@{\extracolsep{5pt}}ccc@{}}
				\hline
				\multirow{2}{*}{\textbf{Method}}  & \multicolumn{2}{c}{\textbf{CLIP score}} \\
    
				
				\cline{2-3} 
				
				
				& Text sim$\uparrow$& Consistency$\uparrow$ \\
				
				\hline
				CSD  &0.1344&0.9541\\
				Gen-1  &0.2340&0.8801\\
				P2V  &0.2271&0.9084\\
                \hline
				
				\textbf{Ours}  & {0.2284}&{0.9473}\\
				\hline
				
			\end{tabular}}
   \subcaption{Video editing}\label{table:video}
\end{minipage}
\captionof{table}{\textbf{Quantitative Evaluation Results}. (a) For 3D Scene editing, our model outperforms the baselines in CLIP scores, indicating  better text alignment and view consistency. (b) In panorama editing, our model shows higher CLIP score with decent LPIPS score, indicating  better text alignment with comparable structural consistency. (c) For video editing, our model outperforms the baselines in CLIP scores, which shows our method shows better text matching and frame consistency. }
\vspace*{-0.4cm}
\end{table}

\noindent\textbf{Quantitative Results. } To quantitatively evaluate our proposed method, we again used CLIP text score to measure output-text alignment. Different from our 3D case, our source image patches are not similar in structural aspect, therefore measure cross-view consistency is not relevant in this case. Instead, we measured the LPIPS perceptual similarity between the cropped patches from source and output images to measure the structural consistency between source and output. For evaluation, we experimented with 5 different images with 5 different text conditions for each image. Table \ref{table:panorama} shows the results. Our model shows best in CLIP text score, and similar score of structural similarity between CSD. Since our baseline Syncdiffusion fails to contain source image structure, it shows degraded performance in both of CLIP and structural score. 

        
			
    
				
				
				
				
				
				

\vspace{-0.2cm}
\subsection{Video Editing}
\vspace{-0.2cm}
\noindent\textbf{Baselines. } For evaluation of video editing, we use three different baseline methods. As recent method which use pre-trained video diffusion model we use open-source editing method of Gen-1~\cite{gen1}. More recent approach of Pix2Vid~\cite{pix2vid} use only 2D diffusion model to sample the context-consistent outputs using whole self-attention feature injection. We also use CSD again for video editing baseline.  

\noindent\textbf{Qualitative Results. } For qualitative results, we show the sampled outputs in Fig.~\ref{fig:fig_video}. For consistent editing methods of CSD, the method often failed to properly edit the output with target text conditions. For video-diffusion based Gen-1, the output follow the text condition but the method suffer from low cross-frame consistency and unwanted artifacts. In Pix2Vid, the output shows better cross frame consistency with proper text-guided semantic change, but still the method suffer from inconsistency when the frame is largely changed from starting point. In our case, we can edit the consecutive images with text condition with consistent context. More samples are in our project page.
    
\noindent\textbf{Quantitative Results. } To quantitatively evaluate the video editing performance, we used same evaluation metric to our 3D editing case. In Table \ref{table:video}, baseline CSD shows lowest text matching score while highest frame consistency, which indicate the the method fails to edit the video. In Gen-1, the method showes highest text alignment score but lowest cross-frame consistency, which also indicate that it fails to generate consistent images. Our method shows best performance considering both of text alignment and frame consistency. 
        
			
    
				
				
				
				
				
				
    
    \subsection{Additional Experiments}
    \vspace{-0.1cm}
\noindent\textbf{Editing with Custom Concept.}
Since our method can accurate apply the text-to-2D image generation model performance to other modality, we can incorporate custom concept-aware editing to our framework. First, we trained the custom concept aware model using 
Custom Diffusion~\cite{custom} framework, then applied our proposed method. Figure \ref{fig:fig_custom} shows the output of custom concept semantic transfer. Our edited outputs follow the overall semantic attribute of custom concepts, while each frames maintain common context.

\noindent\textbf{Editing for Local Regions}
By incorporating additional mask information, we enable localized editing within specific regions of the image. To achieve this, we employ the CLIPseg~\cite{clipseg} model to extract mask information $M_i$ from sequential images $I_i$ using words (e.g. "ground") as queries for the target region. During the sampling process, we combine the U-Net output $\epsilon_i^t$ and DDIM inverted noise $\epsilon^{inv}_i$ through the mask such as: $\epsilon=M_i\cdot\epsilon_i^t+(1-M_i)\cdot\epsilon^{inv}_i$. As demonstrated in the Figure \ref{fig:fig_mask}, this approach allows us to preserve the background while focusing the editing on the desired region. Furthermore, semantic consistency is maintained across images, even within the localized region. 

\textbf{More Details.} Experimental settings, ablation study, and more samples are in our appendix.

\vspace{-0.2cm}
\section{Conclusion}
\vspace{-0.2cm}
In this paper, we present a novel editing method that enables unified editing across diverse modalities, including 3D scenes, videos, and panorama images, utilizing only a basic 2D image text-to-image diffusion model. Our approach leverages the sequential nature of images in different modalities and introduces a novel inverted feature injection technique to preserve context information while maintaining high editing performance. Extensive experiments demonstrate the superior editing capabilities of our method compared to existing baseline models on 3D scenes, videos, and panorama.


\small
\bibliographystyle{ieeenat_fullname}






\newpage
\appendix

\section*{Appendix}
\section{Implementation Details}
\noindent\textbf{3D Scene Editing.} For 3D scene editing, we utilized the dataset provided by original NeRF~\cite{nerf}. Specifically, we conducted experiments on the local light field fusion (LLFF) dataset, which consists of real-world scenes. The LLFF dataset comprises 20-40 multi-view images for a single scene along with corresponding camera geometry information. We were able to reconstruct edited 3D scenes by sequentially editing training images using our proposed sampling method for a scene, then trained the NeRF model on the edited images. For this purpose, we set the reference image $I_{ref}$ as the first image in training set $I_{0}$ and set the remaining training images as sequential images. All images were edited using Stable Diffusion v2.1 and have a resolution of 512x512. In all cases, we used $T=50$ for timestep, and for $t_{edit}$, we set it to 20 when significant semantic changes were required and 15 for style transfer where the structure was largely preserved. $t_{context}$ was set to 50. We employed the TensoRF~\cite{tensorf} model as the NeRF backbone model. Instead of training from scratch using the edited scenes, we fine-tuned for approximately 5,000 steps from a 3D scene model pre-trained with the source images. During the training process, we used the same hyperparameters setting to the previous work.

\noindent\textbf{Panorama Image Editing.} For panorama image editing, we referred to the source code of MultiDiffusion~\cite{multidiffusion}. The images were resized to a resolution of 512x2048, and during training, the images were cropped into sub-images with a stride of 256 and a patch size of 512. We used series of cropped images as the editing image set. In these images, we employed sequential referencing strategy which is using  the previous sequence image $I_{i-1}$ as reference image. More specifically, for the editing process of $I_1$, $I_0$ served as the reference $I_{ref}$, and for $I_2$ editing, $I_1$ became the reference. We employed MultiDiffusion's sampling method which is averaging the score estimation of the overlapping region by the amount of overlapping times, enabling seamless editing of high-resolution images. For timestep, we also used $T=50$. In this case, we set both $t_{edit}$ and $t_{context}$ to 20, as good sample quality could be obtained by not injecting attention for later timesteps.

\noindent\textbf{Video Editing.} For video editing, we utilized the DAVIS dataset~\cite{davis}. We extracted frames from the video and edited them individually. Since editing full frames took a significant amount of sampling time, we subsampled approximately 30-40 frames for editing. Similar to the panorama image case, we employed a sequential sampling approach where the previous frame served as the reference for editing the subsequent frame. For timestep, we used $T=50$ and $t_{edit}$ was set to 15. $t_{context}$ was set to 50.

\noindent\textbf{Efficiency Comparison.}
During the experiments, we conducted all the experiments using single RTX 3090 24GB GPU. Since our method require sequential image editing, sampling time varies by different modalities. In 3D scene editing, overall sampling time for 35 training images is 12 minutes. For NeRF fine-tuning with edited images, 5,000 step takes about 3 minutes.  The baseline ED-NeRF which require training additional latent NeRF takes about 14 minutes. Instruct Nerf2Nerf takes longer time over 1 hour as it requires alternative dataset change. For CLIP-based editing methods of CLIP-NeRF and NerF-art, the method require 6 minutes and 15 minutes, respectively. Considering the output quality, our proposed method has better efficiency compared to baseline methods.

For panorama image editing, the cropped images are 7 samples if we use 512x2048 image. Therefore the sampling time is about 3 minutes. In case of baseline CSD, it takes about 2 minutes for 200 step optimization. In case of syncdiffusion, the method also takes about 3 minutes.

In video editing, we usually used 40 frames which takes about 14 minutes in our case. In Pix2Vid, the method takes about 12 minutes as it use similar sequential sampling process. In CSD, the method takes longer time over 20 minutes, as it require optimization process for all frames. In Gen-1, sampling time is about 1 minute, as we used faster cloud GPU service which is provided by the authors. Overall, out method shows similar efficiency compared to baseline methods when it comes to training time and GPU resources.

\section{Ablation Study}
\begin{figure}[t!]
    \centering
    \includegraphics[width=1.0\linewidth]{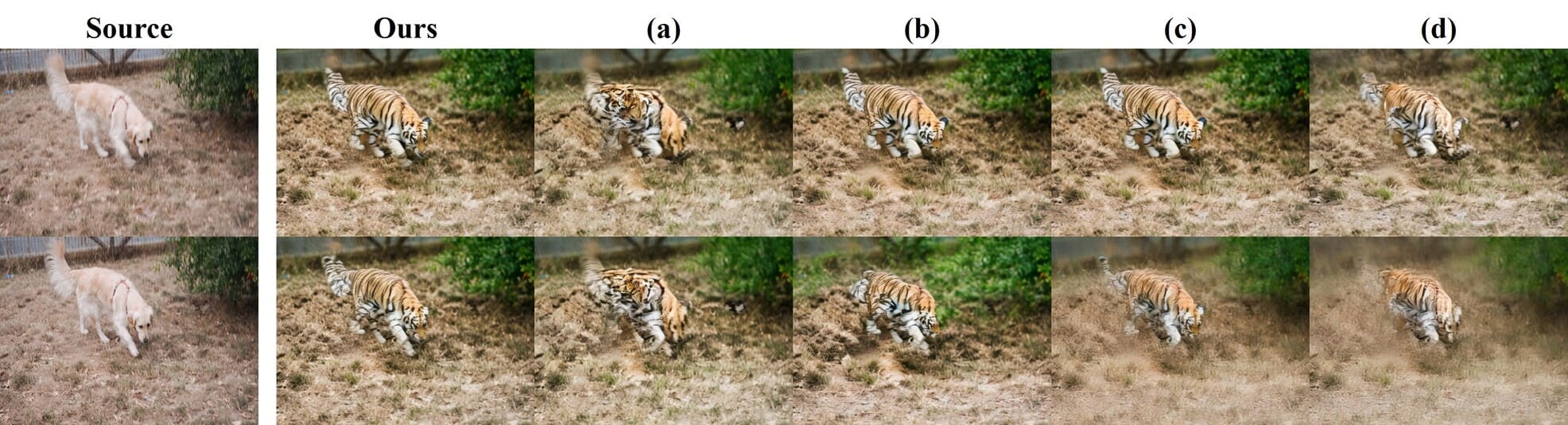}
    \caption{\textbf{Ablation Study Results.} Results for editing video using text prompt \textit{"tiger"}. Our best setting successfully edit the source to target text semantic while preserving the context. (a) without injecting inverted features of resnet, overall context semantic is not maintained. (b) Without injecting inverted query features, the output loses context consistency in detail. (c) Without injecting inverted key features, the output structure is severely damaged. (d) Without injecting inverted value features, the output context is not maintained from reference. 
}
    \label{fig:fig_ablation}
    \end{figure}

To assess the significance of each component in our inverted feature injection technique, we conducted an ablation study by evaluating different attention injection scenarios against our best setting. A shown in Fig.~\ref{fig:fig_ablation},
(a) when we excluded resnet features during inverted feature injection, the overall structure of the output images was partially preserved. However, the images deviated significantly from the core content of the source image, often transforming into unrelated images. This highlights the crucial role of residual layer features in maintaining the core content of the edited images. (b) Eliminating query features from the feature injection stage resulted in the inability of the sampled output to follow the structure of the source image at the detail level. This indicates that query features play a vital role in capturing fine-grained details and ensuring that the editing process aligns with the source image's structure.
(c) Removing key features from the feature injection stage led to the collapse of the content structure in the output with generating blurry outputs. This demonstrates the essential contribution of key features in conveying contextual information, which is crucial for maintaining the overall structure of the edited image.
(d) Excluding value features from the injection stage also resulted in a significant deterioration of the content structure, leaving only a rough outline of the original image. This suggests that value features are important in preserving the detailed content information.

\section{Limitations and Potential Impacts}
Our method relies on the semantic information of existing frames. Therefore, maintaining consistency can be challenging when the semantic distance between sequential frames is significantly large. This could limit the applicability of our method to long video editing tasks. The ability to edit using obscene text prompts raises ethical concerns, particularly when applied to images of real people. To address these concerns, implementing appropriate language filtering mechanisms is crucial.

\section{Additional Results}
\begin{figure}[t!]
    \centering
    \includegraphics[width=1.0\linewidth]{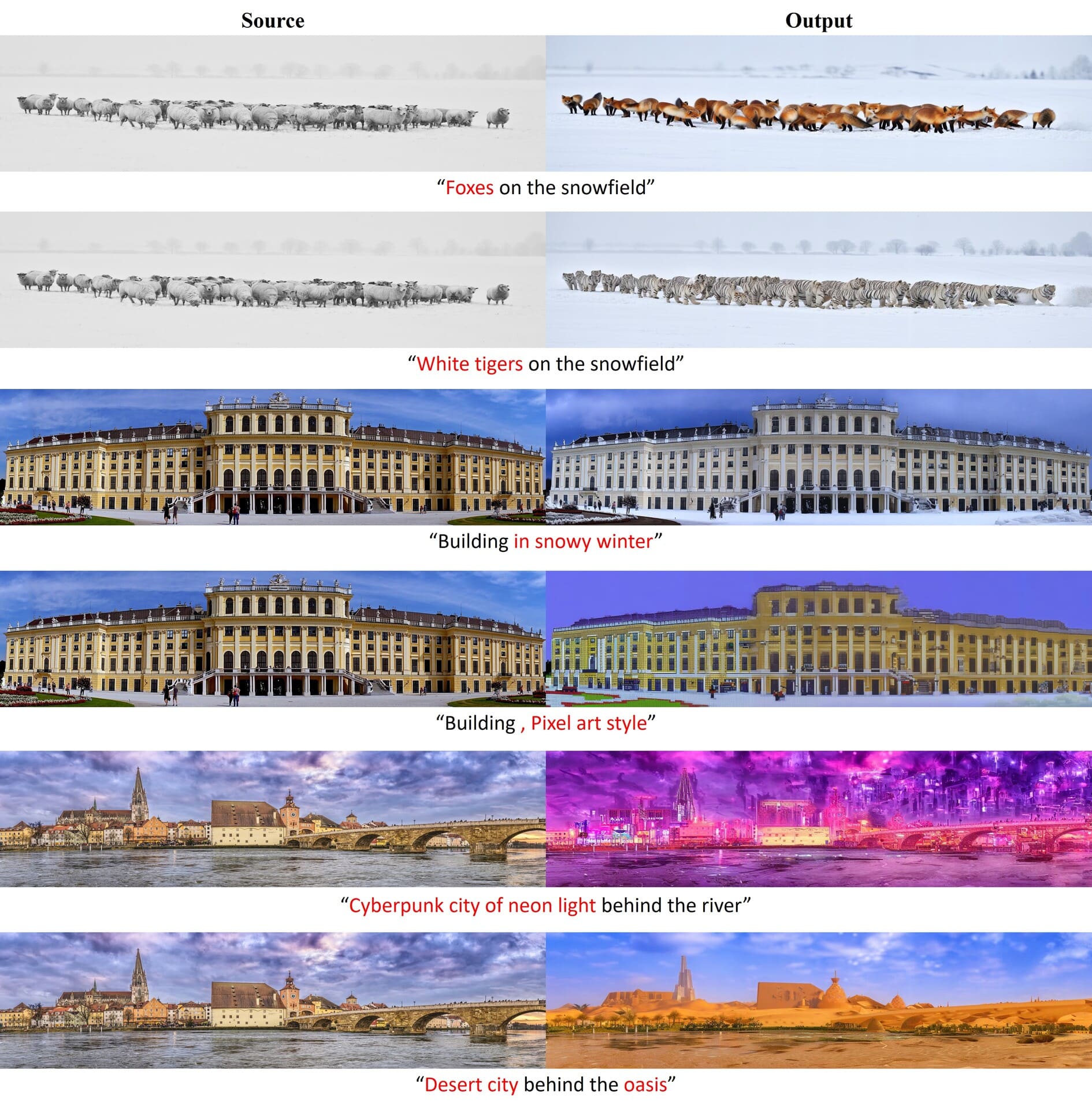}
    \caption{\textbf{Additional Results of panorama image editing.} We show the edited panorama image output.
}
    \label{fig:fig_add_panorama}
    \end{figure}

\begin{figure}[t!]
    \centering
    \includegraphics[width=1.0\linewidth]{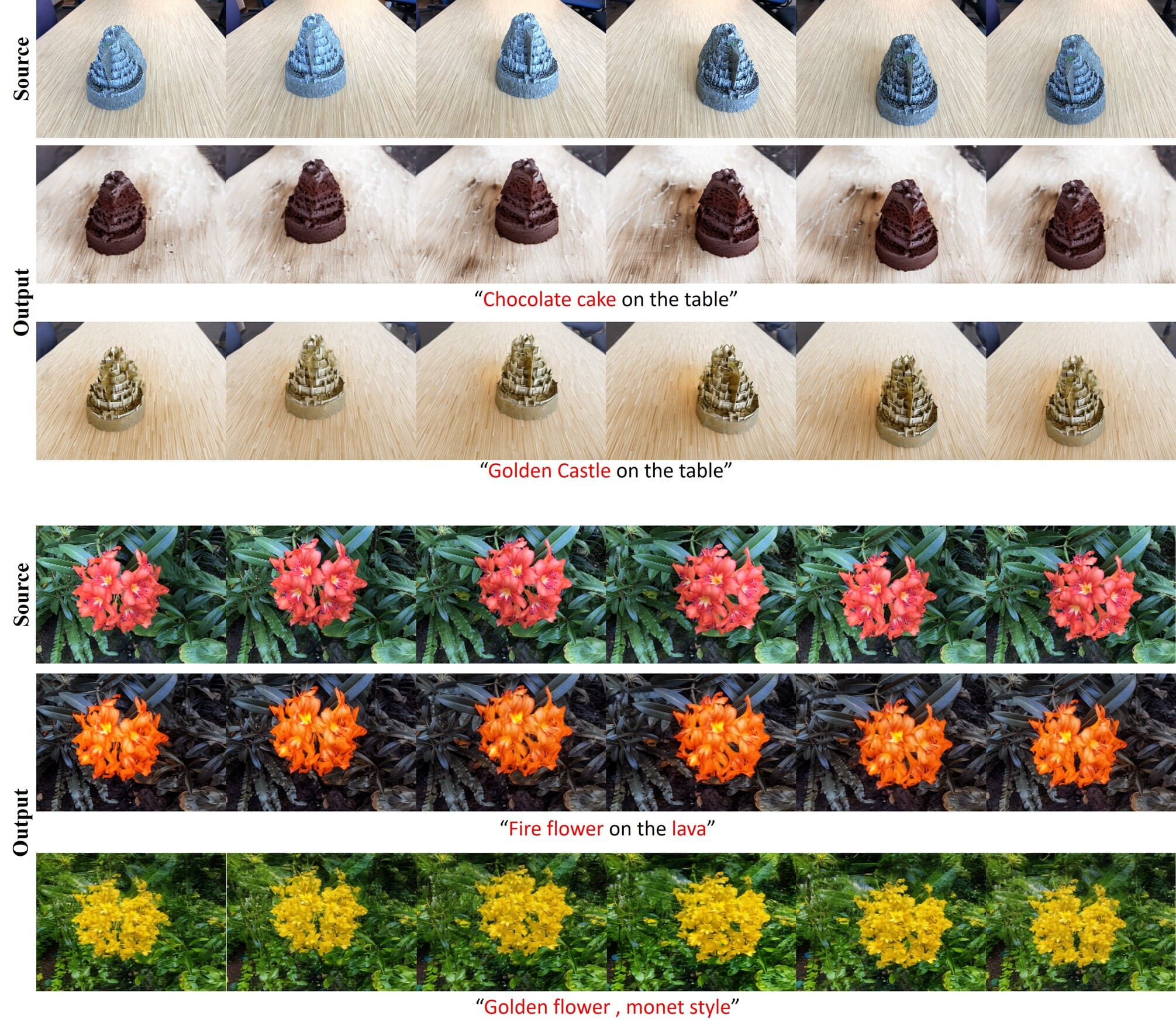}
    \caption{\textbf{Additional Results of 3D scene editing.} We show the edited 3D scene outputs.
}
    \label{fig:fig_add_nerf}
    \end{figure}

\begin{figure}[t!]
    \centering
    \includegraphics[width=1.0\linewidth]{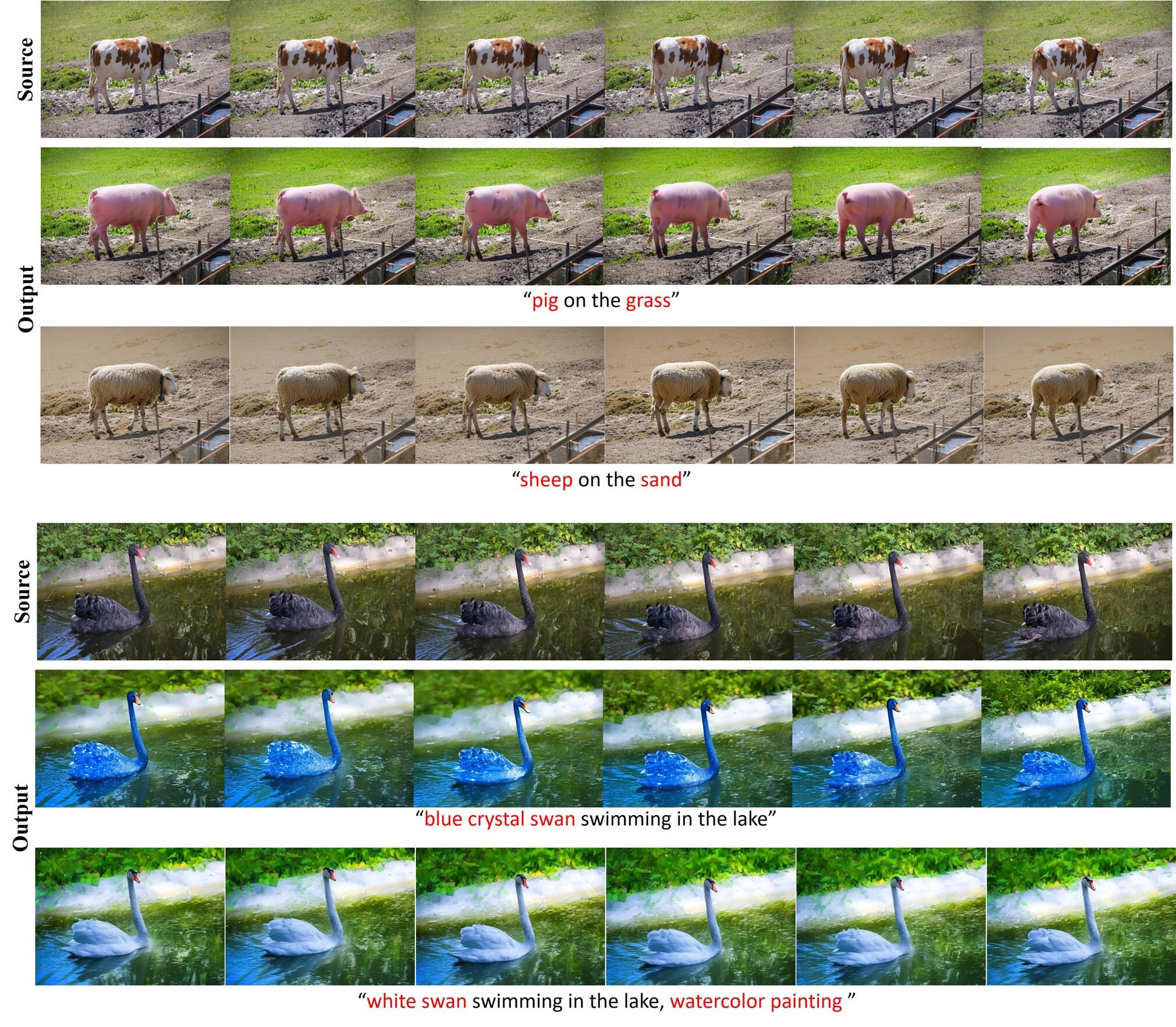}
    \caption{\textbf{Additional Results of video editing.} We show the edited video outputs.
}
    \label{fig:fig_add_video}
    \end{figure}

We show additional results in Figures \ref{fig:fig_add_panorama}, \ref{fig:fig_add_nerf}, and \ref{fig:fig_add_video}
\clearpage

\clearpage

\end{document}